\documentclass[10pt,twocolumn,letterpaper]{article}

\usepackage[pagenumbers]{style} 
\makeatletter
\@namedef{ver@everyshi.sty}{}
\makeatother

\usepackage{graphicx}
\usepackage{amsmath}
\usepackage{amssymb}
\usepackage{booktabs}
\usepackage{tikz, pgfplots}
\usepackage{cuted}

\usetikzlibrary{external}
\usepgfplotslibrary[external]
\pgfplotsset{width=7cm,compat=1.8}
\tikzset{
    error band/.style={fill=orange},
    error band style/.style={
        error band/.append style=#1
    }
}

\newcommand{\ours}{DiffPose}

\usepackage[pagebackref,breaklinks,colorlinks]{hyperref}
\usepackage{xcolor}

\usepackage[capitalize]{cleveref}
\crefname{section}{Sec.}{Secs.}
\Crefname{section}{Section}{Sections}
\Crefname{table}{Table}{Tables}
\crefname{table}{Tab.}{Tabs.}

\newcommand{\first}[1]{\textbf{#1}}

\newcommand{\parag}[1]{{\bf{#1}}}

\newcommand{\vc}{\mathbf{c}}

\newcommand{\ve}{\mathbf{e}}

\newcommand{\vx}{\mathbf{x}}

\newcommand{\mI}{\mathbf{I}}

\usepackage{amsmath,amsfonts,bm}

\def\eqref#1{equation~\ref{#1}}

\def\1{\bm{1}}

\def\eps{{\epsilon}}

\def\vc{{\bm{c}}}

\def\ve{{\bm{e}}}

\def\vx{{\bm{x}}}

\def\evx{{x}}

\def\mI{{\bm{I}}}

\DeclareMathAlphabet{\mathsfit}{\encodingdefault}{\sfdefault}{m}{sl}
\SetMathAlphabet{\mathsfit}{bold}{\encodingdefault}{\sfdefault}{bx}{n}

\newcommand{\E}{\mathbb{E}}

\begin{document}

\title{DiffPose: Multi-hypothesis Human Pose Estimation using Diffusion Models}

\author{Karl Holmquist \qquad Bastian Wandt\\
Linköping University\\
{\tt\small [name.surname]@liu.se}
}
\maketitle

\begin{abstract}
Traditionally, monocular 3D human pose estimation employs a machine learning model to predict the most likely 3D pose for a given input image.
However, a single image can be highly ambiguous and induces multiple plausible solutions for the 2D-3D lifting step which results in overly confident 3D pose predictors.
To this end, we propose \emph{\ours}, a conditional diffusion model, that predicts multiple hypotheses 
for a given input image. 
In comparison to similar approaches, our diffusion model is straightforward and avoids intensive hyperparameter tuning, complex network structures, mode collapse, and unstable training.
Moreover, we tackle a problem of the common two-step approach that first estimates a distribution of 2D joint locations via joint-wise heatmaps and consecutively approximates them based on first- or second-moment statistics.
Since such a simplification of the heatmaps removes valid information about possibly correct, though labeled unlikely, joint locations, we propose to represent the heatmaps as a set of 2D joint candidate samples.
To extract information about the original distribution from these samples we introduce our \emph{embedding transformer} that conditions the diffusion model.
Experimentally, we show that DiffPose slightly improves upon the state of the art for multi-hypothesis pose estimation for simple poses and outperforms it by a large margin for highly ambiguous poses.

\end{abstract}

\section{Introduction}
\label{sec:intro}
\begin{figure}
    \centering
    \resizebox{\linewidth}{!}{%
    \includegraphics[width=0.49\textwidth]{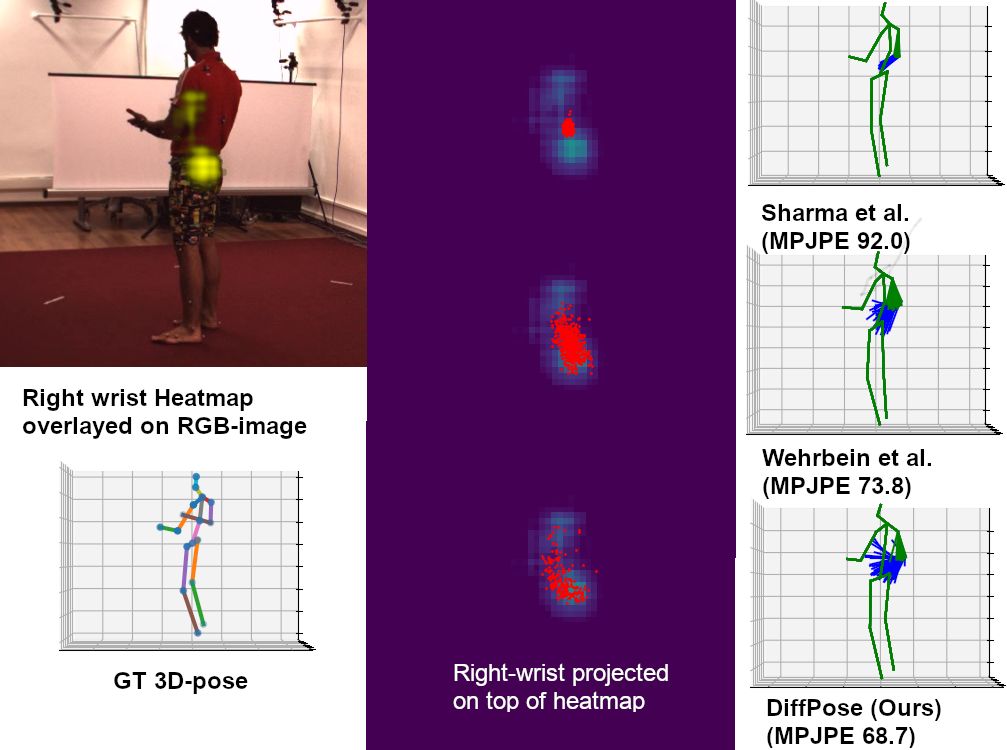}%
    }
    \caption{Comparison of our approach to Sharma~\etal\cite{Sharma_2019_ICCV} and Wehrbein~\etal\cite{wehrbein2021probabilistic}. While \cite{Sharma_2019_ICCV} produces very similar poses, even for uncertain detections, \cite{wehrbein2021probabilistic} achieves a higher diversity. However, they oversimplify heatmaps as a Gaussian and, thus, struggle with different uncertainty distributions. Note that the densest region of samples (red) for \cite{Sharma_2019_ICCV} and \cite{wehrbein2021probabilistic} is very similar and at a point with low certainty. By contrast, \ours\ produces 3D poses that cover the full uncertainty in the heatmap leading to a lower error.}
    \label{fig:teaser}
\end{figure}
Human pose estimation from monocular images is an open research question in computer vision with many applications, \eg in human-machine interaction, autonomous driving, animation, sports, and medicine.
Recent advances in deep learning based human pose estimation show promising results on the path to highly accurate 3D reconstructions from single images. 
Commonly, a neural network is trained to reconstruct the most likely 3D pose given an input image.
However, the projection from 3D to a 2D plane, which is performed by a camera capturing a person, results in an inevitable loss of information.
This lost information cannot be uniquely reconstructed and, therefore, we argue that a meaningful 3D human pose estimator must be able to recover the full distribution of possible 3D poses for a given 2D pose, \eg as a set of poses with different likelihoods. %
Moreover, downstream tasks can be built to benefit from unlikely poses, for example, consider an autonomous vehicle making decisions based on a single output versus being able to see all possible, though unlikely, outcomes.
Consequently, the interest in this field, called multi-hypothesis human pose estimation, is rising \cite{Jahangiri2017GeneratingMD,Li_2019_CVPR,oikarinen2020graphmdn,Sharma_2019_ICCV,kolotouros2021probabilistic,wehrbein2021probabilistic,li2022mhformer}.
Some approaches estimate a small fixed-size set of poses \cite{Jahangiri2017GeneratingMD,Li_2019_CVPR,oikarinen2020graphmdn,li2022mhformer} which are not able to fully represent the real output distribution.
Others are based on variational autoencoders \cite{Sharma_2019_ICCV} or normalizing flows \cite{kolotouros2021probabilistic,wehrbein2021probabilistic} and can predict an infinite set of poses that provides a stronger approximation for the 3D pose distribution.
However, they require complex architectures and lack diversity in their outputs since the 2D input data is simplified as shown in \cref{fig:teaser}.

Our goal is a multi-hypothesis human pose estimator that is easy to train and produces high-quality samples covering the full range of possible and plausible output poses.
To this end, we make three major contributions: we 1) are the first to represent a 3D human pose distribution with a conditional diffusion model which, in its surprisingly simple architecture and training, achieves state-of-the-art results, 2) use the full 2D input information from heatmaps without any simplifications by our novel sampling strategy, and 3) propose a transformer architecture that handles these samples without losing information about joint uncertainties.

Neural diffusion models recently gained huge interest due to their impressive performance in image generation \cite{ramesh2022hierarchical,rombach2022high,saharia2022photorealistic}.
We exploit their capability to generate even subtle details that formerly were only achievable by hard-to-train GANs \cite{Wandt2019RepNet,davydov2022adversarial} or normalizing flows \cite{xu2020ghum,kolotouros2021probabilistic,wehrbein2021probabilistic,wandt2022elepose}.
Even in its simplicity, our diffusion model creates meaningful human poses and unlike VAEs and GANs, it does not suffer from mode collapse, posterior collapse, vanishing gradients, and training instability \cite{Kobyzev2020nfreview}.
While pose representations via normalizing flows also do not show such phenomena they require a sophisticated model of the human kinematic chain \cite{xu2020ghum}, a kinematic chain prior \cite{wehrbein2021probabilistic}, and additional care during training.
By contrast, our diffusion model is robust during training and creates meaningful poses without requiring further constraints. 

Our second major contribution reveals a problem in current two-step approaches that first predict 2D joint positions in an image and consecutively use these predictions as input to the 3D reconstruction step.
While this enables the 3D estimator to be agnostic to the input image and consequently promises generalization across image domains, it removes valid structural and depth information that can only be seen in the images.
We exploit that almost all 2D human pose detectors employ heatmaps encoding joint occurrence probabilities as an intermediate representation.
Traditionally, the maximum argument of these heatmaps is used as input to the second stage which removes all information about the uncertainty of the detector.
Few approaches extract additional information, such as confidence values \cite{WanRud2021a} or Gaussian distributions fitted to the heatmap \cite{wehrbein2021probabilistic}.
However, they still oversimplify the heatmap as shown in \cref{fig:teaser}, therefore, missing important details.
To this end, we propose to condition the diffusion model with an embedding vector computed from a set of joint positions directly sampled from the heatmaps.
We build a so-called \emph{embedding transformer} which combines joint-wise samples and their respective confidences to a single embedding vector that encodes the distribution of the joints.

Our code and trained models will be released upon acceptance.

\section{Related Work}
Monocular 3D human pose estimation is a huge field with vast and diverse approaches. 
Hence, this section focuses on the closest related work, namely 2-stage approaches\footnote{A 2D joint detection step is followed by a 3D lifting step.} and competing multi-hypothesis methods.
In contrast to approaches that estimate a 3D human body shape \cite{multi_bodies_biggs2020,KanazawaCVPR18,kocabas2019vibe,kolotouros2019spin,li21hybrik,pavlaCVPR18,xu2020ghum,Zanfir2020WeaklyS3}, we focus on predicting the 3D locations of a set of predefined joints.

\parag{Lifting 2D to 3D.}
We follow the vast body of work that estimates 3D poses from the output of a 2D pose detector
\cite{Moreno-Noguer_2017_CVPR,Chen_2017_CVPR,Ci_2019_ICCV,fang2018learning,inthewild3d_2019,Hossain2018ECCV,Li_2020_CVPR,PhysCapTOG2020,Wandt2019RepNet,WanRud2021a,DBLP:journals/corr/abs-1905-07862,Xu_2020_CVPR}. 
These two-stage approaches decouple the difficult problem of 3D depth estimation from the easier 2D pose localization.
With the 3D lifting step being agnostic to the image data it is easily transferable to other image domains, e.g. in-the-wild data. %
Moreover, in contrast to 3D training data, 2D images are significantly easier to annotate and, therefore, a huge amount of labeled in-the-wild images is already readily available which reduces bias towards indoor scenes that are common in 3D datasets.
Early work in learning-based pose estimation is done by Akhter and Black~\cite{Akhter_2015_CVPR} who learn a pose-conditioned joint angle limit prior to restrict invalid 3D pose reconstructions. 
The simplest and very influential approach that commonly serves as a baseline is proposed by Martinez \etal~\cite{martinez_2017_3dbaseline}, who employ a fully-connected residual network to lift 2D detections to 3D poses, surprisingly outperforming previous approaches by a large margin.

The approaches above predict a single most likely pose for a given input.
By contrast, we predict a set of plausible 3D poses 
from a single 2D pose.
Additionally, we leverage the full output heatmap of the 2D pose detector which formerly was simplified to it's maximum, an uncertainty label \cite{BogoECCV2016,WanRud2021a,Xu_2020_CVPR}, or Gaussian distributions \cite{wehrbein2021probabilistic}.
With our novel heatmap sampling strategy we are able to reflect the full uncertainty of the 2D predictor in our 3D pose hypotheses.

\parag{Multi-hypothesis 3D human pose estimation.}
Ambiguities of monocular 3D human pose estimation and sampling multiple 3D poses via heuristics is discussed in early work \cite{LeeCo04,Simo12,SmiTri01,SmiTri03}.
Recently, few approaches are proposed that use generative machine learning models which generate multiple diverse hypotheses to cover the ambiguous nature of 3D human pose estimation.
Jahangiri and Yuille~\cite{Jahangiri2017GeneratingMD} uniformly sample from learned occupancy matrices \cite{Akhter_2015_CVPR} to generate multiple hypotheses from a predicted seed 3D pose.
They use a rejection sampling approach based on a 2D reprojection error in combination with bone lengths constraints.
Li and Lee~\cite{Li_2019_CVPR} learn the multimodal posterior distribution using a mixture density network (MDN) \cite{370fbeadb5584ba9ab2938431fc4f140}.
They define a 3D hypothesis by the conditional mean of each Gaussian kernel.
Oikarinen \etal~\cite{oikarinen2020graphmdn} improve \cite{Li_2019_CVPR} by utilizing the semantic graph neural network of \cite{zhaoCVPR19semantic}.
A major limitation is the requirement of an a priori decided number of hypothesis. %
By contrast, Sharma \etal~\cite{Sharma_2019_ICCV} condition a variational autoencoder with 2D pose detections which is able to produce an unlimited amount of hypotheses.
They rank the 3D pose samples by estimated joint-ordinal depth relations from the image. 
Kolotouros \etal \cite{kolotouros2021probabilistic} estimate parameters of the SMPL body model \cite{loper2015smpl} using a conditional normalizing flow.
Wehrbein \etal~\cite{wehrbein2021probabilistic} also propose a normalizing flow to model the posterior distribution of 3D poses. 
They stabilize the training by a multitude of losses including a pose discriminator network similar to generative adversarial networks \cite{goodfellow2020generative}.
By contrast, our diffusion-based 3D pose estimator requires only a single loss and converges stably while improving upon previous approaches, especially on a selected subset of very ambiguous poses.
Moreover, we show that our approach generates more physically plausible poses.
Unlike \cite{wehrbein2021probabilistic} we do not simplify the 2D heatmaps as a Gaussian distribution but instead leverage the entire heatmap enabling 3D pose predictions that fully reflect the uncertainty in the 2D predictions.
Li \etal \cite{li2022mhformer} employ a transformer to learn a distribution from temporal data that is represented by 3 hypotheses which are later merged to predict a single one.
In contrast, \ours\ can predict an infinite amount of poses, therefore, representing the distribution more accurately and does not require temporal data.

\section{Method}
\begin{figure*}
    \centering
    \resizebox{\linewidth}{!}{%
    \includegraphics{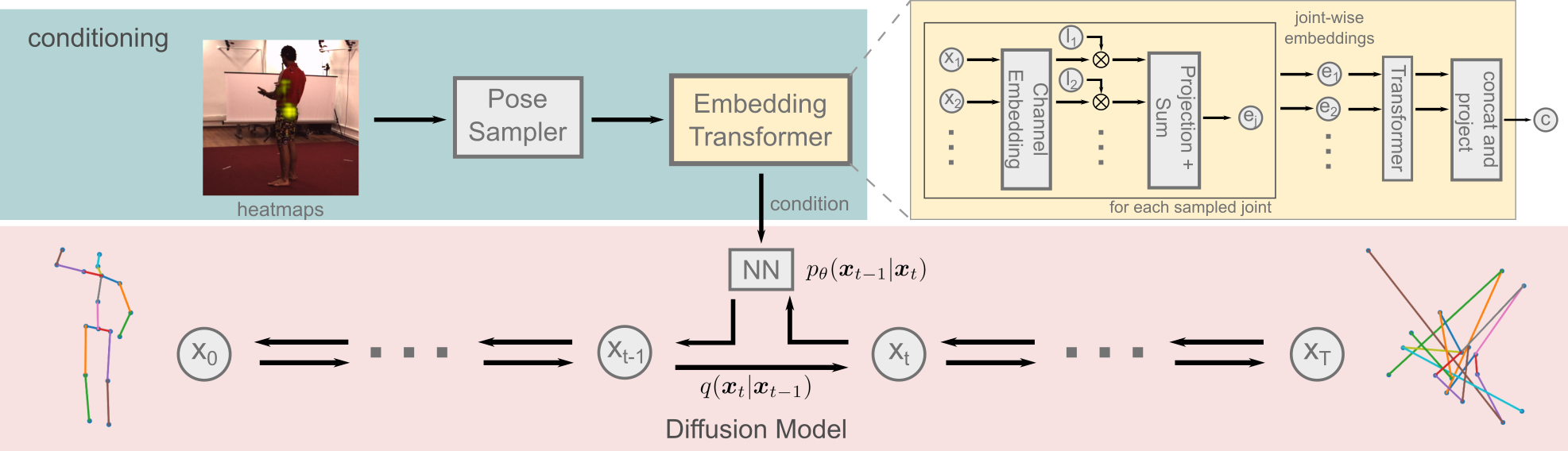}
    }%
    \caption{Overview of our proposed method. It consists of two parts: the diffusion model and the conditioning. The diffusion model alone is able to generate meaningful 3D poses. To generate multiple hypotheses for the 2D to 3D lifting process the conditioning on the 2D heatmaps in each step of the denoising process  is a crucial part. Using our proposed heatmap sampling in combination with our embedding transformer that predicts an embedding for all sampled poses we achieve diverse and meaningful 3D pose predictions.}
    \label{fig:diffusion_model}
\end{figure*}

Our aim is to generate realistic and accurate 3D human poses which approximate the full posterior distribution by utilizing a generative model.
Similar to normalizing flows, which have previously been used for multi-hypothesis pose generation \cite{wehrbein2021probabilistic}, we model the ambiguity caused by the loss of information when projecting 3D data into the image plane by conditioning a diffusion model on the 2D detections. %
Our model is inspired by Denoising Diffusion Probabilistic Models (DDPMs) \cite{ho2020denoising} because of their recent impressive performance and stable training in image generation compared to previous generative models.
We formulate the diffusion process as the iterative distortion of a vector containing 3D joint coordinates 
into a Gaussian distribution $\mathcal{N}(0, \mI)$.
The denoising process is conditioned on %
joint-wise heatmaps that are generated by the 2D joint detector HRNet \cite{Sun_2019_CVPR} using our embedding transformer.
\cref{fig:diffusion_model} shows the full model.

Our second major contribution provides a solution for the problem that all previous two-stage approaches suffer from, namely the loss of valuable uncertainty information when mapping from heatmaps to joint positions.
Previous work has primarily either utilized the maximum likelihood estimate from the 2D joint detector, included confidence values for individual joints \cite{WanRud2021a}, or fitted a Gaussian to approximate the heatmaps \cite{wehrbein2021probabilistic}. 
However, while the heatmap for simpler poses without occlusions can be well represented as a Gaussian, it can be misleading for more complex situations, \eg heatmaps with multi-modal distributions that often occur for occluded joints as shown in Fig.~\ref{fig:teaser}.
As such, we directly utilize the predicted joint position likelihoods to sample the heatmap and utilize both the samples themselves as well as their individual likelihoods to condition the reverse diffusion process.

\subsection{Diffusion Model}
The diffusion model consists of two parts, each defined as a Markov chain: 1) \textit{the forward process} which iteratively adds Gaussian noise of pre-defined mean and variance to the original data, gradually distorting the data and 2) \textit{the reverse process} which is performed by a neural network trained on a step-wise version of the degradation.

\paragraph{The forward process}
is the approximate posterior $q(\vx_{1:T}|\vx_0)$ modeled by a Markov chain that gradually adds Gaussian noise to the original data $\vx_0$ to transform it to a Gaussian distribution $\mathcal{N}(0,\mI)$.
It is performed by a pre-defined noise schedule which adds noise parameterized by $\beta_t$ depending on the step $t$, to the original signal $\vx_0$.
We adopt the cosine-schedule proposed by \cite{nichol2021improved}, which adds a smaller amount of noise near $t=0$ compared to a linear schedule.
At each step $t$ the noise is incrementally added to the signal according to
\begin{equation}
    q(\vx_t | \vx_{t-1}) := \mathcal{N}(\vx_t; \sqrt{1-\beta_t}\vx_{t-1}, \beta_t\mI).
\end{equation}
This formulation allows for sampling of degraded samples at any given time-step in closed form by
\begin{equation}
    q(\vx_t | \vx_{0}) := \mathcal{N}\left(\vx_t; \sqrt{\Bar{\alpha}_t}\vx_{0}, (1 - \Bar{\alpha}_t)\mI\right),
\end{equation}
where $\alpha_t := 1-\beta_t$ and $\Bar{\alpha}_t := \prod_{s=1}^t \alpha_s$.

\paragraph{The reverse process} is the joint distribution $p_\theta(\vx_{0:T})$ and iteratively reverts the degradation by estimating a Gaussian distribution,
\begin{equation}
    p_\theta(\vx_{t-1}|\vx_t) := \mathcal{N}(\vx_{t-1}; \bm{\mu}_\theta(\vx_t, t, \vc), \Sigma_\theta(\vx_t, t)) .
\end{equation}
We follow DDPM by setting $\Sigma_\theta(\vx_t, t) = \beta_t \frac{1-\Bar{\alpha}_{t-1}}{1-\Bar{\alpha}_{t}} \mI$ and parameterize the predicted mean in terms of the current data $\vx_t$ and the predicted noise $\epsilon_\theta$ conditioned on $\vc$,
\begin{equation}
    \bm{\mu}_\theta(\vx_t, t, \vc) = \frac{1}{\sqrt{\alpha_t}}\left( \vx_t - \frac{\beta_t}{\sqrt{1 - \Bar{\alpha}_t}} \epsilon_\theta(\vx_t, t, \vc) \right) .
\end{equation}
For a derivation and more details we refer the reader to \cite{ho2020denoising}.

The noise is predicted using a neural network, parameterized by $\theta$, that takes as input a single input vector build by concatenating the preprocessed conditioning vector $\vc$ which is an embedding of the joint-wise heatmaps, the current 3D pose $\vx_t$ and the current time-step $t$. 
Details about the construction of the condition vector and the exact network architecture are described in \cref{subsect:Sampling} and \cref{sec:implementation_details}, respectively.

\subsection{Sampling from Heatmaps}
In order to represent the heatmaps in a compact yet concise way, we interpret it as an independent, multinomial distribution over possible detections in a $64\times 64$-grid (the output dimension of each heatmap from HRNet \cite{Sun_2019_CVPR}) and draw $n$ samples with replacement for each joint.
For uncertain joints, \eg when they are occluded, the distribution can be highly asymmetric and previous methods struggle to approximate them as shown in \cref{fig:teaser}.
The sampled 2D poses are normalized such that the heatmap covers the interval $[-1, 1]$ in both image directions.
In addition to the sampled poses, we include the most likely 2D pose as one of the samples. 

\subsection{Conditioning the Diffusion Model}
\label{subsect:Sampling}
While integrating a condition into a diffusion model
can be done in many ways\cite{tashiro2021csdi,sinha2021d2c}, there are two key aspects that need to be considered:
1) the individual joint heatmaps are independent, \ie they do not contain any cross-correlation between the joints, %
and 2) directly averaging individual joint samples will result in a loss of the multi-modal information contained in them.
We address both with our \textit{embedding transformer} which is split into two steps as illustrated in \cref{fig:diffusion_model}.
In a first step, we embed all samples for each joint non-linearly into a single vector, thus, maintaining their multi-modality.
Subsequently, to account for inter-joint relationships, these embeddings are used as the input for a transformer network.

\paragraph{The joint-wise embedding}
needs to contain the positional information of each sample as well as its respective likelihood given by the heatmap value.
We use \emph{channel embeddings} to create a non-linear embedding which maintains high spatial-precision while assuring far off positions result in orthogonal embeddings \cite{felsberg2015unbiased}. %
The channel embedding is a soft-histogram of $K$ evenly spaced bins, which uses a predefined basis function to accumulate the samples instead of the rectangular non-overlapping basis of standard histograms. 
We utilize the truncated $\cos^2$-basis,
\begin{equation}
    b(x) = 
    \begin{cases}
        \cos^2(\frac{\pi x}{h})     & \textrm{for }|x|<\frac{h}{2} \\
        0                           & \text{else} ,
    \end{cases} 
\end{equation}
where the bandwidth is $h = \frac{8}{K}$, to let each basis accumulate information from all samples within a distance of 4 bins from the center location.
The channel embedding is first applied to each spatial dimension independently to create a non-linear embedding which is then concatenated to a single vector per sample.
Each embedding is scaled by the likelihoods of the corresponding individual joint samples which forces the following steps to not simply ignore it.
The scaled embedding is passed through a linear layer to introduce sample-wise spatial cross-dependencies.
Finally, the individual joint samples are combined into a single joint embedding $\ve_j$ with
\begin{equation}
    \ve_j = \sum_{n=0}^N \text{NN}\left(l^n  \left[
    \begin{array}{c}
        b\left(\evx_x^n + \frac{2s}{K}\right) \\
        b\left(\evx_y^n + \frac{2s}{K}\right)
    \end{array} \right]_{s=-\frac{K}{2}}^{\frac{K}{2}}\right)
    ,
\end{equation}
where $l^n$ is the likelihood for the sampled joint.

\paragraph{Inter-joint dependencies}
are introduced by adding learned positional encodings to the embeddings in order to distinguish the joints and passing them to a transformer network.
The outputs of the transformer are joint-wise embeddings which can now also contain information about other joints.
To create the final combined conditioning vector $\vc$ the embeddings are concatenated and passed through a linear projection layer.

\subsection{Implementation Details}
\label{sec:implementation_details}

\paragraph{Optimization.}
Both the denoiser and the conditioning are optimized jointly by minimizing the simplified loss objective from Ho \etal \cite{ho2020denoising} 
\begin{equation}
    \mathcal{L} := \E_{t, \vx_0, \epsilon} \left[ \lVert \epsilon - \epsilon_\theta(\sqrt{\Bar{\alpha}_t}\vx_0 + \sqrt{1 - \Bar{\alpha}_t}\epsilon, t, \vc) \rVert^2 \right]
\end{equation}
sampled at uniform time steps, $t$, where $\eps \sim \mathcal{N}(0, I)$.
In contrast to previous work in multi-hypothesis 3D human pose estimation, we only require a single loss term and one neural network which makes training simple and stable for a wide range of hyperparameters.

\paragraph{The denoiser} is a linear layer followed by two residual blocks, each containing two fully connected layers of dimension 1024 with a LeakyReLU as activation function, similar to \cite{martinez_2017_3dbaseline}.
For inference efficiency, our proposed method only samples the heatmap and calculates the condition vector once per forward-pass instead of generating new samples at each time-step $t$.

\paragraph{2D detector.} We use the state-of-the-art and publicly available model HRNet \cite{Sun_2019_CVPR}, pretrained on MPII \cite{andriluka14cvpr} and fine-tuned on Human3.6M \cite{wehrbein2021probabilistic}. The model is trained to predict a heatmap of a standard Gaussian with $\sigma_\text{gt} = 2$px centered on the joint position. 
While any model that produces a heatmap of each individual joint would be possible to use, we chose this one specifically for comparability with previous methods \cite{wehrbein2021probabilistic}.

\paragraph{Data preprocessing.} The raw $64\times64$-pixel heatmaps are generated from cropped square regions as in \cite{wehrbein2021probabilistic}. The sampled 2D joint positions are normalized to the range [-1,1]. 
The 3D poses are processed in decimeters and mean centered individually.

\paragraph{Training.}
The network is trained for 700k iterations using Adam \cite{Adam_KingmaB14}, a learning rate of $1\times10^{-4}$, and a batch size of 64.
We set $K=64$ for the channel embedding and project the final condition into a $2\times 64 \times J=2048$-dimensional vector before concatenating it with the time step $t$ and $\vx_{t-1}$.
During training we randomly drop individual joints by setting the joint embedding, $\ve_j = 0$, with a fixed probability of $0.01$. We noticed that this further improves the symmetry of generated poses and decreases the PA-MPJPE on H36MA (cf. \cref{tab:ablation}).
The training on a single NVIDIA A40 takes approximately 7 hours.

\section{Experiments}
Following previous work, we evaluate our method on the well-known benchmark datasets Human3.6M \cite{h36m_pami} and MPI-INF-3DHP \cite{mono-3dhp2017} using their established training and test splits.
For the Human3.6M dataset, we follow standard protocols and evaluate on every 64th frame of the test set.

Since our main focus are highly ambiguous poses we evaluate on the H36MA subset of Human3.6M as defined by Wehrbein \etal \cite{wehrbein2021probabilistic}.
It contains samples where at least one Gaussian that is fitted to the heatmaps has a standard deviation larger than 5 px.
This subset contains 6.4\% of all samples present in the Human3.6M test set.
These samples are extremely challenging since the joint detector gives inaccurate or wrong results.
The results on this dataset can be seen as the main target %
of our approach.

In addition to Human3.6M and MPI-INF-3DHP, we use the Leeds Sports Pose extended (LSPe) dataset \cite{Johnson11} for qualitative evaluation.

\paragraph{Metrics.}
\label{sec:metrics}
For Human3.6M we follow the standard protocols.
Protocol I calculates the mean Euclidean distance between the root-aligned reconstructed poses and ground truth joint coordinates which is commonly known as \textit{mean per joint position error} (MPJPE).
Protocol II first employs a Procrustes alignment between the poses before calculating the MPJPE, also known as PA-MPJPE.
For 3DHP we additionally report the \textit{Percentage of Correct Keypoints} (PCK).
It is the percentage of predicted joints that are within a distance of $150mm$ or lower to their corresponding ground truth joint.
Following Wandt et al. \cite{WanRud2021a} we additionally evaluate the Correct Poses Score (CPS) which, unlike the PCK, classifies a pose as correct if all joints of the pose are correctly estimated for a given threshold, therefore, yielding a stronger metric than the PCK.
To be independent of a threshold value, the CPS calculates the area under the curve in a range from $0mm$ to $300mm$.
For comparability to prior work we report the performance of the best model for all metrics unless stated otherwise.
Results over multiple runs with their mean and standard deviation are reported in the supplemental document.

\subsection{Quantitative Evaluation}
We report metrics for the best 3D pose hypothesis generated by our network which is in line with previous work.
This evaluation reflects how well the learned 3D poses cover the actual ground truth distribution, which is particularly interesting for ambiguous examples.
Therefore, instead of validating whether predictions are equal to a specific solution, we evaluate if that specific solution is contained in the set of predictions.
Additionally, we evaluate the predicted pose when sampling from the mean during the denoising step (denoted as $z_0$) which roughly corresponds to the most likely pose learned by the diffusion model.

\paragraph{Evaluation on Human3.6M.}
Following \cite{Sharma_2019_ICCV} and subsequent work, we produce $M = 200$ hypotheses for each 2D input. 
Table~\ref{tab:h36m} compares our approach to others and shows that we slightly improve upon the state of the art.
Note that we almost match Li \etal \cite{li2022mhformer} in MPJPE and even outperform them by $7\%$ in PA-MPJPE although they use temporal data.

However, our main target are highly ambiguous cases.
Therefore, our core result is the evaluation on H36MA, the hard subset of Human3.6M, which is shown in Table~\ref{tab:h36ma}. In average, we significantly outperform the state of the art by $6.4mm$ ($9\%$) and $6.5mm$ ($12\%$) in MPJPE and PA-MPJPE, respectively.
Furthermore, we improve the PCK by $1.2\%$ and the CPS by $25.7$.
Figure~\ref{fig:teaser} shows an example of the increased diversity that results in predictions being closer to the ground truth which leads to these large improvements.

\begin{table*}[]
    \centering
    \caption{Results in millimeters for the H36M dataset for protocol 1 (MPJPE) and protocol 2 (PA-MPJPE). The row marked with dagger $\dagger$ uses temporal information and is included for conciseness but not marked in bold even if it shows the best performance for some activities.}
    \resizebox{0.98\linewidth}{!}{%
    \begin{tabular}{ccccccccccccccccc}
    Protocol 1 (MPJPE)  &   Direct. &   Disc.   &   Eat &   Greet   &   Phone   &   Photo   &   Pose    &   Purch.  &   Sit &   SitD    &   Smoke   &   Wait    &   WalkD   &   Walk    &   WalkT   &   Avg. \\ \hline
    Martinez \etal \cite{martinez_2017_3dbaseline} ($M = 1$)  & 51.8 & 56.2 & 58.1 & 59.0 & 69.5 & 78.4 & 55.2 & 58.1  & 74.0 & 94.6 & 62.3 & 59.1 & 65.1 & 49.5 & 52.4 & 62.9 \\ 
    Li \etal~\cite{li2020weakly} ($M=10$)&{62.0} & {69.7} & {64.3} & {73.6} & {75.1} & {84.8} & {68.7} & {75.0} & {81.2} & {104.3} & {70.2} & {72.0} & {75.0} & {67.0} & {69.0} & {73.9} \\
    Li \etal~\cite{Li_2019_CVPR} ($M=5$) & 43.8 & 48.6 & 49.1 & 49.8 & 57.6 & 61.5 & 45.9 & 48.3 & 62.0 & 73.4 & 54.8 & 50.6 & 56.0 & 43.4 & 45.5 & 52.7\\
    Oikarinen \etal \cite{oikarinen2020graphmdn} ($M=200$)&40.0 & {43.2} & {41.0} & {43.4} & {50.0} & {53.6} & {40.1} & {41.4} & {52.6} & 67.3 & {48.1} & \first{44.2} & \first{44.9} & 39.5 & {40.2} & {46.2}\\
    Sharma \etal \cite{Sharma_2019_ICCV} ($M=10$)&$\first{37.8}$ & {43.2} & 43.0 & 44.3 & 51.1 & {57.0} & {39.7} & {43.0} & {56.3} & {64.0} & {48.1} & {45.4} & {50.4} & {37.9} & {39.9} & 46.8\\
    $\dagger$MHFormer Li \etal \cite{li2022mhformer} ($M=3$) & 39.2 & 43.1 & 40.1 & 40.9 & 44.9 & 51.2 & 40.6 & 41.3 & 53.5 & 60.3 & 43.7 & 41.1 & 43.8 & 29.8 & 30.6 & 43.0\\
    Wehrbein \etal \cite{wehrbein2021probabilistic} ($M=1$) & 52.4 & 60.2 & 57.8 & 57.4 & 65.7 & 74.1 & 56.2 & 59.1 & 69.3 & 78.0 & 61.2 & 63.7 & 67.0 & 50.0 & 54.9 & 61.8 \\
    Wehrbein \etal \cite{wehrbein2021probabilistic} ($M=200$)   & 38.5 & \first{42.5} & 39.9 & \first{41.7} & \first{46.5} & 51.6 & 39.9 & 40.8 & \first{49.5} & \first{56.8} & 45.3 & 46.4 & 46.8 & 37.8 & 40.4 & 44.3 \\
    \hline
    Ours ($z_0$) ($M = 1$)                                      
    & 58.7 & 63.4 & 50.7 & 64.5 & 66.7 & 74.6 & 58.7 & 60.9 & 71.1 & 89.5 & 59.5 & 69.6 & 67.5 & 58.2 & 54.2 & 64.5 \\
    Ours ($M = 200$)                                            
    & 38.1 & 43.1 & \first{35.3} & 43.1 & 46.6 & \first{48.2} & \first{39.0} & \first{37.6} & 51.9 & 59.3 & \first{41.7} & 47.6 & 45.4 & \first{37.4} & \first{36.0} & \first{43.3} \\

    \hline \hline
    Protocol 2 (PA-MPJPE)  &   Direct. &   Disc.   &   Eat &   Greet   &   Phone   &   Photo   &   Pose    &   Purch.  &   Sit &   SitD    &   Smoke   &   Wait    &   WalkD   &   Walk    &   WalkT   &   Avg. \\ \hline
    Martinez \etal \cite{martinez_2017_3dbaseline} ($M = 1$)  & 39.5 & 43.2 & 46.4 & 47.0 & 51.0 & 56.0 & 41.4 & 40.6 & 56.5 & 69.4 & 49.2 & 45.0 & 49.5 & 38.0 & 43.1 & 47.7 \\ 
    Li \etal~\cite{li2020weakly} ($M=10$) & {38.5} & {41.7} & {39.6} & {45.2} & {45.8} & {46.5} & {37.8} & {42.7} & {52.4} & {62.9} & {45.3} & {40.9} & {45.3} & {38.6} & {38.4} & {44.3}\\
    Li \etal~\cite{Li_2019_CVPR} ($M=5$) & {35.5} & {39.8} & {41.3} & {42.3} & {46.0} & {48.9} & {36.9} & {37.3} & {51.0} & {60.6} & {44.9} & {40.2} & {44.1} & {33.1} & {36.9} & {42.6}\\
    Oikarinen \etal \cite{oikarinen2020graphmdn} ($M=200$) & 30.8 & 34.7 & {33.6} & {34.2} & {39.6} & {42.2} & {31.0} & 31.9 & {42.9} & 53.5 & {38.1} & {34.1} & {38.0} & {29.6} & {31.1}& {36.3}\\
    *Sharma \etal \cite{Sharma_2019_ICCV}($M=200$) & {30.6} & {34.6} & 35.7 & 36.4 & 41.2 & 43.6 & 31.8 & {31.5} & 46.2 & {49.7} & 39.7 & 35.8 & 39.6 & 29.7& 32.8 & 37.3\\
    $\dagger$MHFormer Li \etal \cite{li2022mhformer} ($M=3$) & 31.5 & 34.9 & 32.8 & 33.6 & 35.3 & 39.6 & 32.0 & 32.2 & 43.5 & 48.7 & 36.4 & 32.6 & 34.3 & 23.9 & 25.1 & 34.4\\
    Wehrbein \etal \cite{wehrbein2021probabilistic} ($M=1$)     & 37.8 & 41.7 & 42.1 & 41.8 & 46.5 & 50.2 & 38.0 & 39.2 & 51.7 & 61.8 & 45.4 & 42.6 & 45.7 & 33.7 & 38.5 & 43.8 \\
    Wehrbein \etal \cite{wehrbein2021probabilistic} ($M=200$)   & \first{27.9} & \first{31.4} & 29.7 & \first{30.2} & 34.9 & 37.1 & \first{27.3} & 28.2 & \first{39.0} & 46.1 & 34.2 & 32.3 & 33.6 & \first{26.1} & 27.5 & 32.4 \\
    \hline
    Ours ($z_0$) ($M = 1$)                                      & 40.0 & 42.4 & 38.5 & 43.8 & 47.4 & 49.5 & 39.7 & 39.2 & 56.7 & 67.6 & 44.7 & 42.7 & 46.3 & 42.3 & 37.6 & 45.2 \\
    Ours ($M = 200$)                                            & 28.1 & 31.5 & \first{28.0} & 30.8 & \first{33.6} & \first{35.3} & 28.5 & \first{27.6} & 40.8 & \first{44.6} & \first{31.8} & \first{32.1} & \first{32.6} & 28.1 & \first{26.8} & \first{32.0} \\
    
    \end{tabular}%
    }
    \label{tab:h36m}
\end{table*}

\begin{table}[]
    \centering
    \caption{Results for the hard subset H36MA as defined by Wehrbein \etal \cite{wehrbein2021probabilistic}. We outperform all comparable methods by a large margin. Additionally, the symmetry error shows that in average \ours~produces more plausible poses.}
    \resizebox{0.98\linewidth}{!}{%
    \begin{tabular}{cccccc}
        \hline
        Method & MPJPE $\downarrow$ & PA-MPJPE $\downarrow$ & PCK $\uparrow$ & CPS $\uparrow$ & Sym $\downarrow$ \\
        \hline
        Li \etal~\cite{Li_2019_CVPR} & 81.1 & 66.0 & 85.7 & 119.9 & - \\
        Sharma \etal~\cite{Sharma_2019_ICCV} & 78.3 & 61.1 & 88.5 & 136.4 & 23.9\\
        Wehrbein \etal~\cite{wehrbein2021probabilistic} & 71.0 & 54.2 & 93.4 & 171.0 & 27.4\\
        \hline
        \ours~(Ours) & $\first{64.6}$ & $\first{47.7}$ & $\first{94.6}$ & $\first{196.7}$ & $\first{14.9}$
    \end{tabular}%
    }
    \label{tab:h36ma}
\end{table}

\paragraph{Generalization to other datasets.}
We evaluate on the MPI-INF-3DHP dataset to show the generalization abilities of our model.
The 2D detector and the diffusion model remain the same as for the Human3.6M dataset and are not trained or refined for the experiments in this section. 
Table~\ref{table:3dhp} shows that, in average, we perform on par with the closest competitor Li \etal \cite{li2020weakly} as shown in the last column.
For the challenging outdoor scenes (column \textit{Outdoor}) our method improves by $4.6\%$ and $1.4\%$ upon \cite{li2020weakly} and \cite{wehrbein2021probabilistic}, respectively.
Similar to the results on H36MA, this highlights that \ours\ is well suited for more complicated scenes.
While we follow previous work and only draw 200 samples an increased number of samples improves the performance significantly, as also shown in \cref{fig:ablation_samples}.
The performance saturates at a PCK of $88.0$ which is a large improvement over our result with 200 samples and state of the art.
Results for other metrics for the 3DHP dataset are reported in the supplemental document.

\begin{table}
\begin{center}
\caption{Quantitative results on MPI-INF-3DHP. We outperform all comparable methods which indicates a good generalizability of our models to other sequences without requiring additional training.}
\resizebox{0.98\linewidth}{!}{%
    \begin{tabular}{c c  c  c  c}
    \hline
    Method & Studio GS $\uparrow$ & Studio no GS $\uparrow$ & Outdoor $\uparrow$ & All PCK $\uparrow$ \\ \hline 
    Li \etal~\cite{li2020weakly} & {86.9} & $\first{86.6}$ & {79.3} & \first{85.0}\\
    Li \etal~\cite{Li_2019_CVPR} & 70.1 & 68.2 & 66.6 & 67.9\\ %
    Wehrbein \etal \cite{wehrbein2021probabilistic} & {86.6} & {82.8} &  {82.5} & {84.3}\\
    \hline
    \ours~(Ours) & \first{87.4}	& 82.9	& \first{83.9}	& 84.9 \\
    \end{tabular}%
    }
    \vspace{-1.5em}
    \label{table:3dhp}
\end{center}
\end{table}

\subsection{Qualitative Evaluation}

\begin{figure*}[t]
    \centering
    \includegraphics[width=\linewidth]{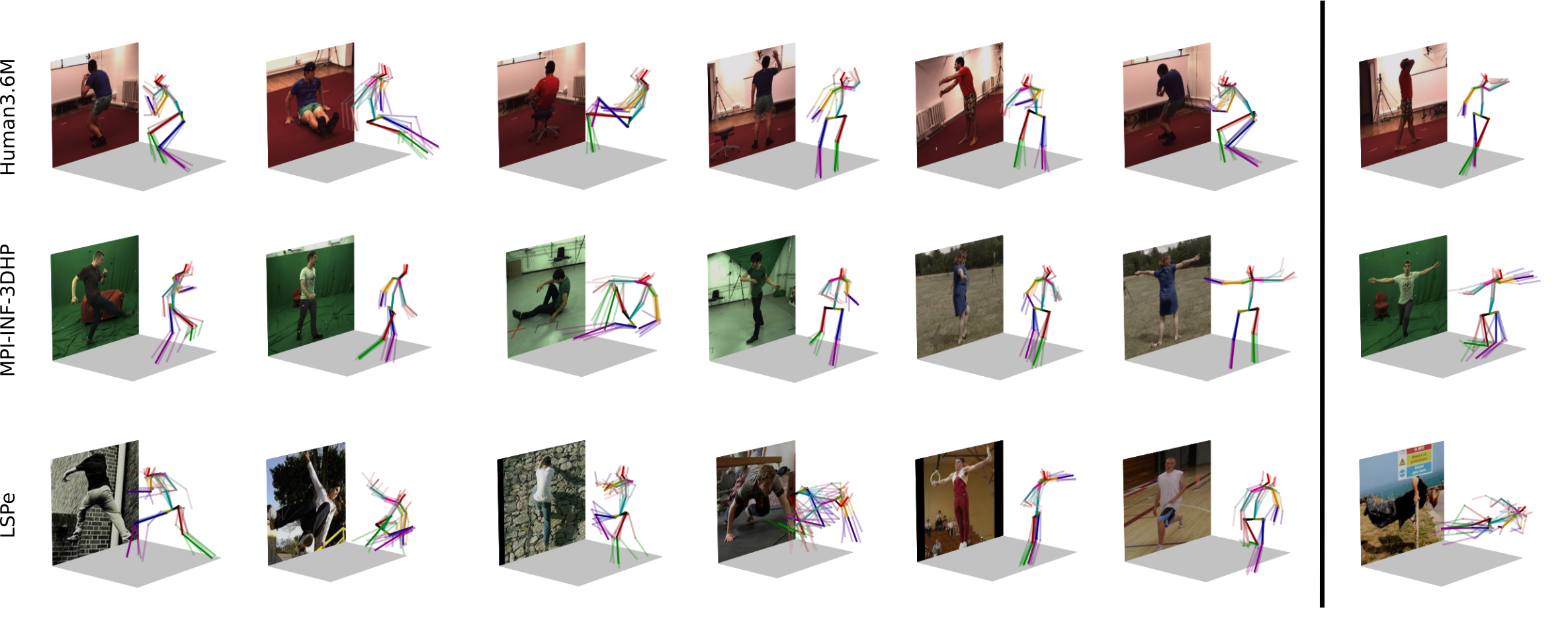}
    \caption{Qualitative results for different datasets. We achieve plausible 3D poses for a large variety of poses. The right most column shows occasional failure cases with misdetected joints (top) and poses far outside the distribution of poses in the training dataset (middle and bottom). For better visibility we only show a subset of the reconstructed poses.}
    \label{fig:qualitative_results}
\end{figure*}

\cref{fig:qualitative_results} shows visual results of our method for 3 different datasets, Human3.6M, MPI-INF-3DHP, and LSPe. 
For better visibility we only show 10 pose samples with the middle one in a stronger color. 
Note that the variance for visible joints in common poses is low whereas rare poses with occluded joints show a high variance in the reconstructions.
Even for MPI-INF-3DHP and LSPe we achieve plausible reconstructions although these datasets were not used for training.
In cases where the reconstructed poses do not completely match the ground truth they still have plausible joint angle limits and bone lengths as also discussed in the ablation studies in \cref{sec:ablation}.
Occasional failure cases occur when joints are misdetected by the 2D joint detector (top, right column), or poses are too far outside of the distribution of the poses in the training dataset (middle and bottom right).

\subsection{Ablation Study}
\label{sec:ablation}
We perform several ablation studies to evaluate our method in different settings and validate our contributions.

\paragraph{Why diffusion models?}
While diffusion models have shown amazing results for highly-detailed image generation little is known about their capabilities to model human skeletons.
To verify that diffusion models are also capable to represent features at a higher abstraction level for humans, we calculate a symmetry error as the mean bone lengths difference between the left and right side of the human body.
Table~\ref{tab:h36ma} shows the results in the column \textit{Sym}. 
Although \cite{wehrbein2021probabilistic} uses a kinematic prior that encourages symmetry we achieve a significantly lower error (12.5mm or 46\%) which means our generated poses are more plausible.
We also outperform our closest competitor \cite{Sharma_2019_ICCV} by $9mm$ or $38\%$.
This is also reflected by the significantly lower PA-MPJPE shown in \cref{tab:h36ma}.

\begin{figure}
\begin{center}
\resizebox{.49\columnwidth}{!}{%
\includegraphics{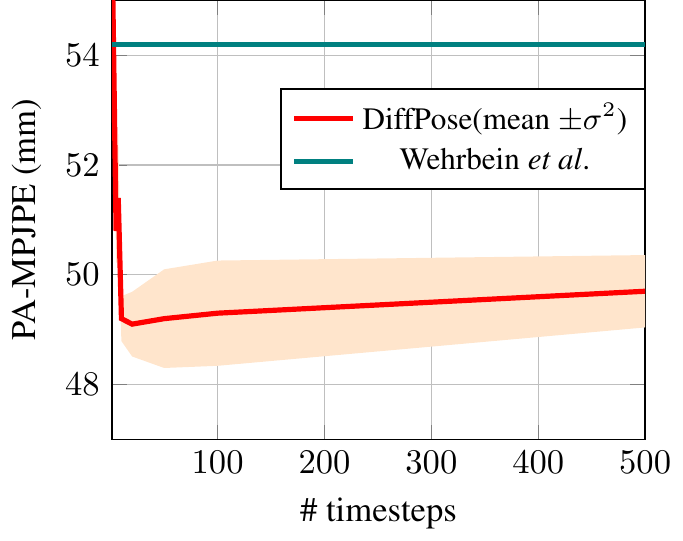}
}
\resizebox{.49\columnwidth}{!}{%
\includegraphics{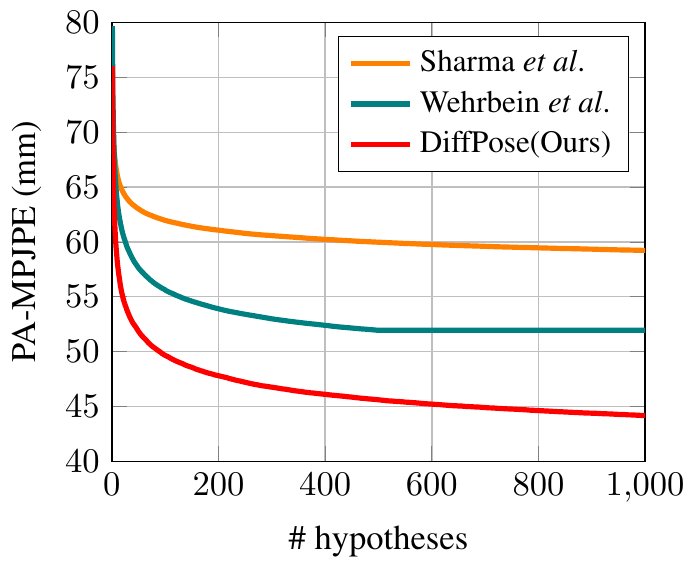}
}
\caption{Evaluation results on the subset H36MA.
\textbf{Left}: increasing number of timesteps in the denoising process. 
The performance saturates at approximately 32 samples. \cite{wehrbein2021probabilistic} is included for comparison. \textbf{Right}: increasing number of hypotheses. Our method continuous to improve, in contrast to \cite{Sharma_2019_ICCV,wehrbein2021probabilistic}.
}
\vspace{-1.5em}
\label{fig:ablation_timesteps_hypos}
\end{center}
\end{figure}

\paragraph{Number of diffusion steps.}
The forward and backward pass in a diffusion model is defined as a Markov process.
To ensure that the forward process results in a Gaussian distribution, infinitely many steps are required.
Commonly, this is approximated with a large finite number of steps.
\cref{fig:ablation_timesteps_hypos} shows the performance of our model for different numbers of total diffusion steps.
Since 25 appears to be the optimal value we perform all our experiments with that number of steps.
A larger value slightly worsens the result but remains relatively constant across a large range of values while still being significantly below our closest competitor.

\paragraph{Embedding transformer.}
\begin{table}[]
    \centering
    \caption{Ablation study for different configurations of \ours. Each of our contributions clearly improves the performance.
    The default setting for the ablation study is 25 timesteps and 24 samples.%
    }
    \resizebox{0.98\linewidth}{!}{%
    \begin{tabular}{lccccc}
        \hline
        Configuration & MPJPE $\downarrow$ & PA-MPJPE $\downarrow$ & PCK $\uparrow$ & CPS $\uparrow$ & Sym $\downarrow$\\
        \hline
        Sample-free denoiser-only model & 75.8 & 55.0 & 91.8 & 169.9 & 18.5 \\
        Sample-based denoiser-only & 73.2          & 52.7      & 92.5  & 177.5     & 21.2 \\
        \hline
        \ours\ w/o sampling            & 67.6          & 49.7          & 93.8          & 191.0             & 15.3 \\
        \hspace{1pt} w/o maximum-likelihood sample       & 68.2          & 49.5          & 94.4          & 193.9             & 14.9 \\
        \hspace{1pt} w/o cross-spatial dependence  & 69.9    & 50.5          & 93.6          & 188.1             & 14.3 \\
        \hspace{1pt} w/o likelihood scaling  & 69.5          & 50.4          & 93.9          & 190.7             & 14.1 \\
        \hspace{1pt} w/o transformer         & 71.5          & 51.9          & 93.3          & 188.9             & 15.5 \\
        \hspace{1pt} w/o dropout             & 70.0          & 50.1          & 93.2          & 187.5             & 16.2 \\
        \hline
        \ours\ (Ours)            & 68.0          & 48.8          & 94.1          & 193.9             & 15.5 \\
    \end{tabular}%
    }
    \label{tab:ablation}
\end{table}

\begin{figure}
\begin{center}
\resizebox{.6\columnwidth}{!}{%
\includegraphics{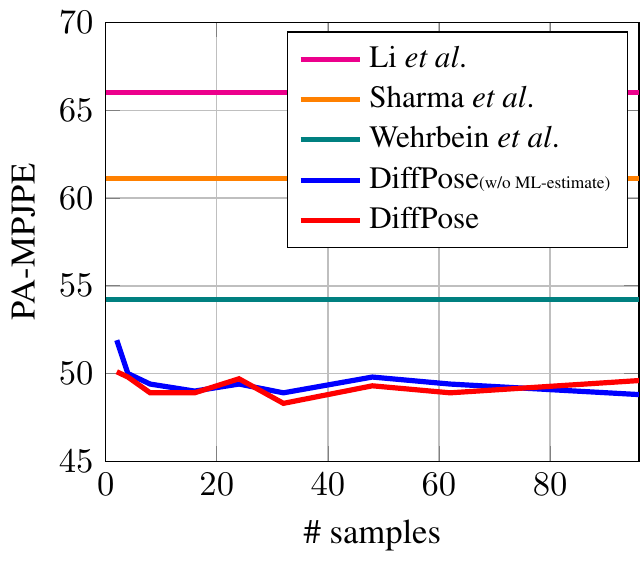}
}
\caption{Evaluation results on the subset H36MA for an increasing number of samples. For comparison, the non-sample based methods \cite{Li_2019_CVPR,Sharma_2019_ICCV,wehrbein2021probabilistic} are included.}
\label{fig:ablation_samples}
\end{center}
\end{figure}

\cref{tab:ablation} shows the performance of our model using different ways to compute the condition vector. 
The possibly simplest condition is directly using the maximum argument of the heatmaps as condition which is shown in row \textit{Sample-free denoiser-only model}.
To represent the heatmap via samples one could also use those directly as condition by ordering them according to likelihood and concatenating them (row \textit{Sample-based denoiser-only}).
However, these simple conditions perform significantly worse compared to our full model.
Our embedding transformer is a more sophisticated way to encode the heatmap information.
The remaining rows show the performance when removing different parts of the embedding transformer.
Each of our proposed individual design choices contributes to the final performance.
While the difference between using our proposed sampling compared to directly using the maximum of the heatmap is not large in \cref{tab:ablation} a clear difference can be seen in the sample diversity in \cref{fig:teaser}.
Excluding the maximum of the heatmap from the sampled poses (row \textit{w/o maximum-likelihood sample}) shows only a minor difference, especially as the number of samples increases, as also visualized in \cref{fig:ablation_samples}.
This underlines that our embedding transformer indeed learns to represent the full heatmap.

\paragraph{Number of samples.}

The number of samples drawn from the heatmap plays a crucial role for the representative power of the embedding created by the embedding transformer.
Fig.~\ref{fig:ablation_samples} shows the performance for different numbers of samples.
While a single sample is not enough, as also shown in \cref{tab:ablation}, the performance increases with more samples.
We choose 32 samples in our main experiments as a good trade-off between performance and complexity.
Note that the performance remains stable over a wide range of values indicating the robustness of our method against different choices of hyperparameters.
In any configuration we already outperform other methods.

\paragraph{Number of hypotheses.}
\cref{fig:ablation_timesteps_hypos} shows the performance on H36MA for an increasing number of hypotheses compared to others.
As expected with more hypotheses the errors decrease.
Notably, ours continues to improve for more than 1000 hypotheses.
For 2000 hypotheses we reach an MPJPE of $58.3mm$ and a PA-MPJPE of $42.9mm$ which is significantly below the results reported in \cref{tab:h36ma}.

\section{Limitations}
In general, all two-step approaches remove image information in favor of being agnostic to the image domain, \eg indoor/outdoor, lighting, and image size.
While we effectively extract more information from the heatmaps as any other two-step approach, still image information is ignored that could possibly be used to further refine results. 
However, directly incorporating it into current pose estimation methods mostly leads to degraded performance.
Therefore, we still strongly advocate two-stage approaches and encourage extracting other valuable features from the images for further research.
Another issue arising from the intermediate heatmap representation are completely wrong 2D joint detections.
Fig.~\ref{fig:qualitative_results} shows that our model is only partially able to correct for these mistakes since it tries to generate plausible poses in terms of joint angle limits and bone lengths by the strong representational power of the diffusion model. 

\section{Conclusion}
We presented \ours, a conditional diffusion model, that estimates multiple hypotheses for 3D human pose estimation from a single image.
Our diffusion model learns plausible human poses, \eg in terms of symmetry, that are valid solutions for a given input image while not only outperforming previous methods by a large margin for highly ambiguous poses but also being simpler and more robust to train using only a single loss term.
Additionally, we propose a novel sampling method from 2D joint heatmaps in combination with a embedding transformer to represent the uncertainties in the heatmaps.
We show that the embeddings predicted by the transformer are superior to simpler embeddings used in prior work.
We hope that our novel embedding method enables future research to use the full information in 2D joint heatmaps.

Our accurate 3D pose estimates have a wide range of applications in downstream tasks, such as 3D pose tracking, multi-view pose estimation, and likelihood estimation for pose forecasting.

\clearpage
\newpage
{\small
\bibliographystyle{ieee_fullname}
\bibliography{literature.bib}
}

\clearpage
\newpage
\appendix

\begin{strip}
\vspace{-20pt}
\begin{center}
\textbf{\Large Supplementary Material for}
\end{center}
\begin{center}\textbf{\Large DiffPose: Multi-hypothesis Human Pose Estimation using Diffusion Models}
\end{center}
\end{strip}

\section{Additional Results}
\begin{figure*}
\begin{center}
\resizebox{0.3\linewidth}{!}{%
\includegraphics{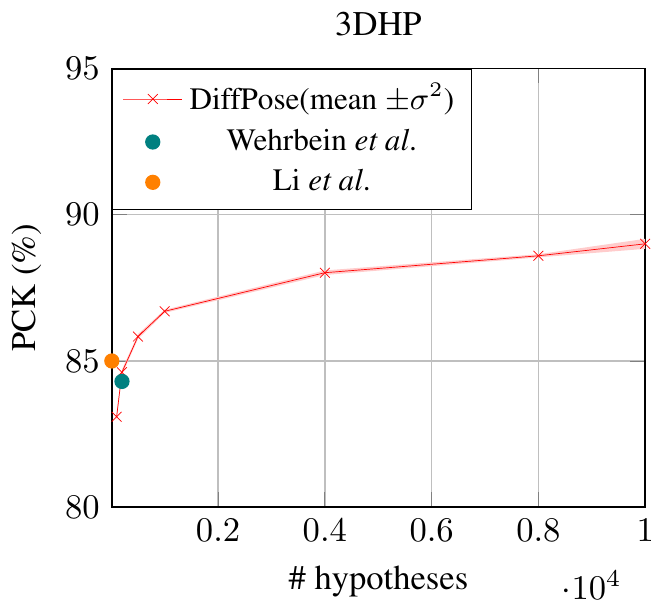}
}%
\resizebox{0.3\linewidth}{!}{%
\includegraphics{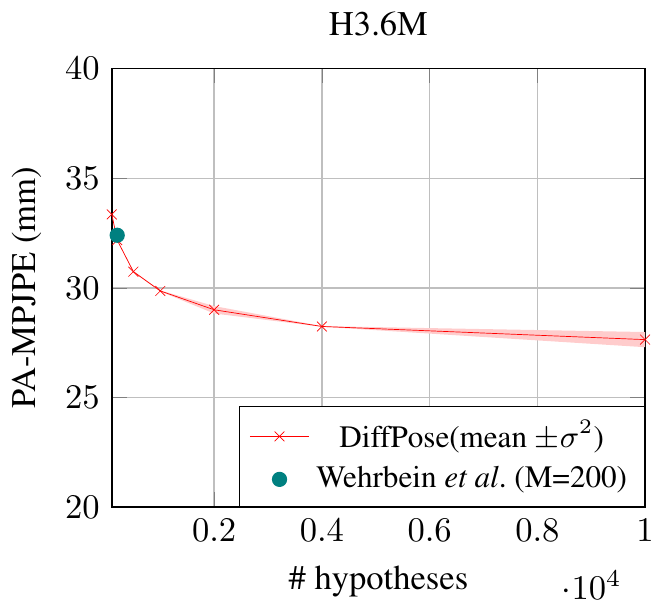}
}%
\resizebox{0.3\linewidth}{!}{%
\includegraphics{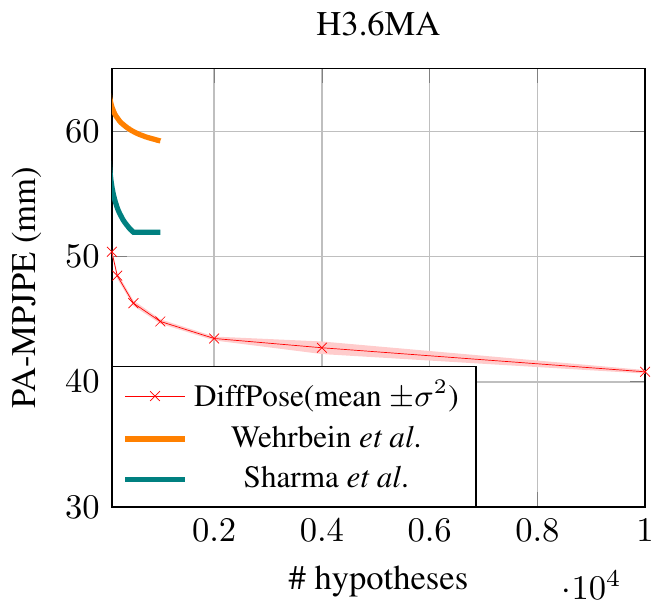}
}
\caption{Plots for the three different datasets illustrating how the performance changes as the number of hypotheses generated increases. The results of \ours is the average of five different models trained with the same architecture and parameter setting, in addition to the mean we also show the variance as a light-red band of $\pm\sigma^2$. However, note that the variance is in general very small and barely visible.}
\label{fig:ksamples}
\end{center}
\end{figure*}

In addition to the experiments, we report results for Human3.6M, H36MA and MPI-INF-3DHP for a larger number of hypothesis in Tables~\ref{tab:sup:h36m}, \ref{tab:sup:h36ma}, and \ref{tab:sup:3dhp}, respectively.
With an increasing number of samples the performance of our model increases significantly.
Fig.~\ref{fig:ksamples} shows this effect visually.
While this behaviour is partially expected our closest competitor \cite{wehrbein2021probabilistic} shows no such improvements and saturates in performance at approximately 500 samples.
This underlines that our model produces more diverse samples that cover the posterior distribution better.

In addition, Tables~\ref{tab:sup:h36m}, \ref{tab:sup:h36ma}, \ref{tab:sup:3dhp}, and \ref{tab:sup:3dhp_other_metrics} show the mean and standard deviation of our model over 5 runs. 
The low standard deviations indicate that our approach consistently achieves good performance.

For completeness we provide results for MPJPE and PA-MPJPE for the MPI-INF-3DHP dataset in Tab.~\ref{tab:sup:3dhp_other_metrics} that were not evaluated by previous methods.

\begin{table}[h]
    \centering
    \caption{Results for the full H36M dataset for different number of hypotheses. Results are reported as mean and variance over 5 runs.}
    \resizebox{0.98\linewidth}{!}{%
    \begin{tabular}{ccc}
        \hline
        Method & MPJPE $\downarrow$ & PA-MPJPE $\downarrow$ \\
        \hline
        Li \etal~\cite{Li_2019_CVPR}~(M=5)                    & 52.7 & 42.6 \\
        Sharma \etal~\cite{Sharma_2019_ICCV}~(M=10)            & 46.8 & 37.3 \\
        Wehrbein \etal~\cite{wehrbein2021probabilistic}~(M=200) & 44.3 & 32.4 \\
        \hline
        \ours~(M=200)                                    & $44.2\pm0.18$ & $32.1\pm0.03$ \\ 
        \ours~(M=500)                                    & $44.0\pm0.14$ & $30.7\pm0.03$ \\ 
        \ours~(M=1000)                                   & $40.7\pm0.19$ & $29.9\pm0.02$ \\ 
        \ours~(M=4000)                                   & $38.3\pm0.17$ & $28.2\pm0.01$ \\ 
        \ours~(M=10000)                                  & $37.4\pm1.40$ & $27.6\pm0.35$ \\ 
    \end{tabular}%
    }
    \label{tab:sup:h36m}
\end{table}

\begin{table}[h]
    \centering
    \caption{Results for the hard subset H36MA as defined by Wehrbein \etal \cite{wehrbein2021probabilistic} for different number of hypotheses. Note that we exclude the symmetry measure in this table since we observed that the mean and variance of $Sym.$ was independent of the number of hypotheses (Sym = $14.9\pm0.02$).}
    \resizebox{0.98\linewidth}{!}{%
    \begin{tabular}{ccccc}
        \hline
        Method & MPJPE $\downarrow$ & PA-MPJPE $\downarrow$ & PCK $\uparrow$ & CPS $\uparrow$ \\ %
        \hline
        Li \etal~\cite{Li_2019_CVPR} & 81.1 & 66.0 & 85.7 & 119.9 \\
        Sharma \etal~\cite{Sharma_2019_ICCV} & 78.3 & 61.1 & 88.5 & 136.4 \\
        Wehrbein \etal~\cite{wehrbein2021probabilistic} & 71.0 & 54.2 & 93.4 & 171.0 \\
        \hline
        \ours~(M=200)   & $66.5\pm1.43$ & $48.5\pm0.23$ & $94.3\pm0.04$ & $194.2\pm2.33$ \\
        \ours~(M=500)   & $63.5\pm1.34$ & $46.3\pm0.21$ & $95.1\pm0.04$ & $201.7\pm1.56$ \\
        \ours~(M=1000)  & $61.6\pm1.15$ & $44.8\pm0.17$ & $95.6\pm0.04$ & $206.6\pm1.49$ \\
        \ours~(M=4000)  & $58.6\pm2.08$ & $42.7\pm0.50$ & $96.2\pm0.08$ & $213.7\pm5.66$ \\
        \ours~(M=10000) & $56.3.5\pm1.19$ & $40.8\pm0.14$ & $96.6\pm0.02$ & $218.9\pm1.44$ \\
    \end{tabular}%
    }
    \label{tab:sup:h36ma}
\end{table}

\begin{table}
\begin{center}
\caption{Quantitative results on MPI-INF-3DHP. This table contains the mean and variance over 5 runs, evaluated using different amount of hypotheses. As can be seen, our method continuous to improve with increasing amounts of hypotheses.}
\resizebox{0.98\linewidth}{!}{%
    \begin{tabular}{c c  c  c  c}
    \hline
    Method & Studio GS $\uparrow$ & Studio no GS $\uparrow$ & Outdoor $\uparrow$ & All PCK $\uparrow$ \\ \hline 
    Li \etal~\cite{li2020weakly} & {86.9} & {86.6} & {79.3} & {85.0}\\
    Li \etal~\cite{Li_2019_CVPR} & 70.1 & 68.2 & 66.6 & 67.9\\ %
    Wehrbein \etal \cite{wehrbein2021probabilistic} & {86.6} & {82.8} &  {82.5} & {84.3}\\
    \hline
    \ours~(M=200) & $87.4\pm 0.35$ &	$82.5 \pm 0.12$ & $83.3 \pm 0.15$	& $84.6 \pm 0.09$ \\
    \ours~(M=500) & $88.5\pm 0.22$ &	$83.9 \pm 0.08$ & $84.4 \pm 0.18$	& $85.8 \pm 0.08$ \\
    \ours~(M=1000) & $89.3\pm 0.22$ &	$84.9 \pm 0.05$ & $85.2 \pm 0.04$	& $86.7 \pm 0.05$ \\
    \ours~(M=4000) & $90.7\pm 0.25$ &	$86.3 \pm 0.04$ & $86.4 \pm 0.14$	& $88.0 \pm 0.08$ \\
    \ours~(M=8000) & $91.2\pm 0.19$ &	$87.0 \pm 0.04$ & $86.8 \pm 0.09$	& $88.6 \pm 0.05$ \\

    \end{tabular}%
    }
    \vspace{-1.5em}
    \label{tab:sup:3dhp}
\end{center}
\end{table}

\begin{table}
\begin{center}
\caption{Quantitative results on MPI-INF-3DHP. This table contains the mean and variance over 5 runs, evaluated using different amount of hypotheses. As can be seen, our method continuously improves with increasing amounts of hypotheses.}
\resizebox{0.6\linewidth}{!}{%
    \begin{tabular}{c cc}
    \hline
    Method & \multicolumn{2}{c}{All} \\ 
    \ours & MPJPE & PA-MPJPE \\
    \hline 
    M=200  & $108.2 \pm 1.69$	& $66.9 \pm 0.45$ \\
    M=500  & $104.3 \pm 2.06$	& $64.4 \pm 0.41$ \\
    M=1000 & $101.5 \pm 2.20$	& $62.8 \pm 0.29$ \\
    M=4000 & $96.7 \pm 1.72$	& $60.0 \pm 0.25$ \\
    M=8000 & $94.5 \pm 1.72$	& $58.6 \pm 0.25$ \\

    \end{tabular}%
    }
    \vspace{-1.5em}
    \label{tab:sup:3dhp_other_metrics}
\end{center}
\end{table}

\section{Network Architecture - details}

\subsection{Condition Embedding}

\paragraph{Positional embedding}
We use 64 basis per dimension to create the positional embedding, the centers are evenly spread on the interval [-1, 1].
The embeddings of the x- and y-coordinate are concatenated into a single vector and passed into a linear layer with 128 input and output channels.
The embedded joint samples are summed into a joint embedding of dimension 128 before a learned positional joint embedding is added to the embedding.
Each individual joint embedding is passed to a transformer encoder with 4 layers, each using 4-heads and a feed-forward dimension of 512.
The modified joint embeddings are concatenated into a single 16$\times$128-dimensional feature vector and projected using a linear layer into a 16$\times$128-dim conditioning vector.

\subsection{Denoiser}
The denoiser concatenates the 128$\times$16-dimensional condition vector, with the 48-dimensional positional vector $\vx$ and the 1-dimensional timestep (in the range [0, \#steps]).
This results in a vector of 16$\cdot$128+48+1=2097-dimensions that is projected into a 1024-dim vector before being processed by two fully-connected ResNet-blocks (w/o any normalization layer).

\subsection{Dropout}
Dropout of joints was used during training by randomly selecting joints with a probability of  1\% and setting the positional embedding  (before the concatenation and the projection-layer) for all samples of those joints to zero.
Multiple or no dropped out joints per pose are possible.

\section{Qualitative examples}
\begin{figure*}
\centering
\resizebox{0.95\linewidth}{!}{%
\includegraphics{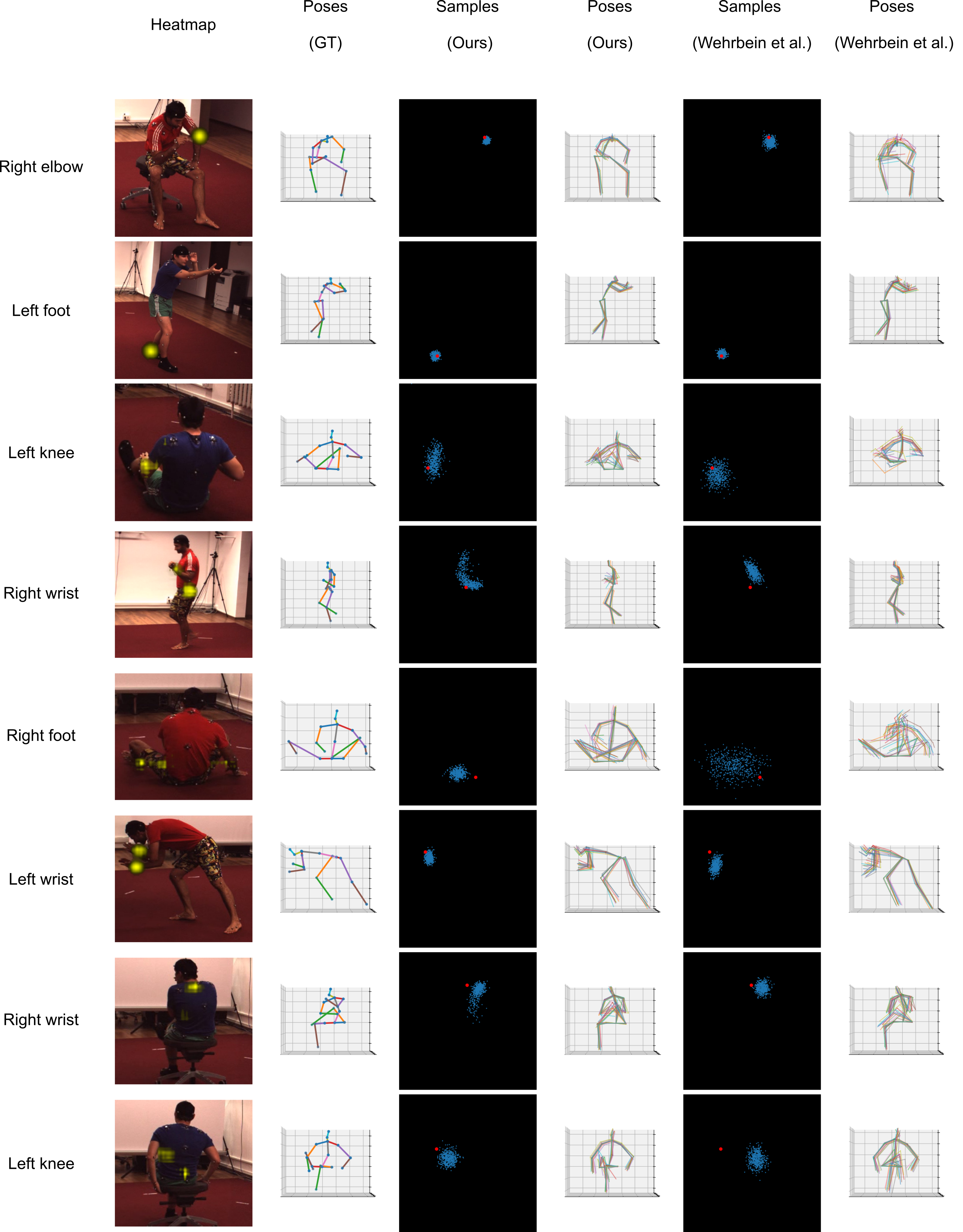}
}
\caption{Qualitative examples for H36MA. For better visibility we only show samples for interesting joints. Our predictions cover the information in the heatmaps well and include the ground truth 3D joint (red dot).}
\label{fig:qualitative}
\end{figure*}

Fig.~\ref{fig:qualitative} shows more qualitative examples.
Joints that are easy to detect result in a very clear heatmap and accordingly a 3D reconstruction with a low diversity (row 1 and 2).
When the 2D joint detector predicts a heatmap with high uncertainty the method of Wehrbein \etal \cite{wehrbein2021probabilistic} struggles to fully cover it.
This leads to 
\begin{enumerate}
    \item massively diverse poses, which are highly implausible (cf. symmetry error in Tab.~2 in the main paper), as shown in row 5, and
    \item over-confident prediction of a set of close poses that might be far away from the ground truth, as shown in rows 4, 6, and 8.
\end{enumerate}
Our method compensates for these effects by either predicting some poses that correspond to lower confidence areas or selecting joint positions that are anatomically plausible.
An example for anatomical plausibility is shown in row 4 where the position of the elbow is clear but the wrist is wrongly detected.
Our model predicts a set of poses where the wrist joint follows an arc which can be interpreted as a rotation around the elbow joint.

\end{document}